# Post-disaster imaged-based damage detection and repair cost estimation of reinforced concrete buildings using dual convolutional neural networks

Xiao Pan[a], T.Y. Yang[b]

[a] PhD candidate, Department of Civil Engineering, University of British Columbia, Vancouver, Canada (p.xiao1994@outlook.com)

[b] Professor, Department of Civil Engineering, University of British Columbia, Vancouver, Canada (yang@civil.ubc.ca)

**Abstract:** *Reinforced concrete buildings are commonly used around the world. With recent earthquakes worldwide, rapid structural damage inspection and repair cost evaluation are crucial for building owners and policy makers to make informed risk management decisions. To improve the efficiency of such inspection, advanced computer vision techniques based on convolution neural networks have been adopted in recent research to rapidly quantify the damage state of structures. In this paper, an advanced object detection neural network, named YOLO-v2, is implemented which achieves 98.2% and 84.5% average precision in training and testing, respectively. The proposed YOLO-v2 is used in combination with the classification neural network, which improves the identification accuracy for critical damage state of reinforced concrete structures by 7.5%. The improved classification procedures allow engineers to rapidly and more accurately quantify the damage states of the structure, and also localize the critical damage features. The identified damage state can then be integrated with the state-of-the-art performance evaluation framework to quantify the financial losses of critical reinforced concrete buildings. The results can be used by the building owners and decision makers to make informed risk management decisions immediately after the strong earthquake shaking. Hence, resources can be allocated rapidly to improve the resiliency of the community.*

## 1 INTRODUCTION

Reinforced concrete (RC) buildings are the most prevalent structural systems constructed worldwide. With many of these buildings built in high seismic zones, the performance of RC building after strong earthquake shaking is becoming a significant concern for many building owners. When an earthquake happens, decision makers such as city planners, emergency management departments need first-hand response to allocate resources to manage the damaged infrastructure. This requires rapid performance assessments of the facilities. Traditional post-earthquake inspections were performed manually and the results may be relatively coarse and highly relying on the proper training of the inspectors and qualitative engineering judgments. The processing time may also be very long, due to a large amount of data processing required. These deficiencies can be overcome if the current manual evaluation processes are fully automated (Zhu & Brilakis, 2010). Research on automation of inspection practices using computer vision-based methods (Jahanshahi, Kelly, Masri, & Sukhatme, 2009; Koch, Georgieva, Kasireddy, Akinci, & Fieguth, 2015; Yeum, & Dyke, 2015; Hoskere, Park, Yoon & Spencer, 2019) has greatly advanced in recent years. In the past, computer vision-based methods were developed based on conventional image processing techniques (IPTs). However, these methods are relatively time-consuming and not robust against background noises. Hence it is ineffective to apply in practice.

Significant achievements have been made in recent years in computer vision with the advancement of artificial neural networks. The development of artificial neural network can be generally divided into 3 phases. The first phase can be dated back to the 1940s–1960s, where the theories of biological learning (McCulloch and Pitts, 1943) and the first artificial neural network such as Perceptron (Rosenblatt, 1958) was implemented. The second phase happened during the 1980–1995 period, where back-propagation technique (Rumelhart, Hinton, & Williams, 1986) was developed to train a neural network with one or two hidden layers. During the 1990s, artificial neural network has evolved to deep neural networks (DNNs), where multiple layers can be trained through back-propagation algorithm. One such application was the work done by LeCun, Bottou, Bengio, & Haffner (1998) for document recognition. The third phase of neural networks (also named deep learning) begun with the breakthrough in 2006 when Hinton, Osindero, and Teh (2006) demonstrated that a so-called deep belief network could be efficiently trained using greedy layer-wise pretraining strategy. With the fast growing and optimization of the deep learning algorithms, the increasing size of training data, as well as enhanced computational power, Convolutional Neural Network (CNN, or ConvNet), which is a class of DNNs, has been advancing rapidly. Unlike traditional neural networks that utilize multiple fully-connected (FC) layers, the hidden layers of a CNN typically include a series of convolutional layers that convolve with a multiplication or other dot product through learnable filters. In recent years, CNNs have dominated the fields of computer vision, speech recognition, and natural language processing.

Within the field of computer vision, CNN such as the AlexNet, developed by Krizhevsky, Sutskever, and Hinton (2012), has shown a substantial increase in accuracy and efficiency than any other algorithms. With the success of AlexNet, CNNs have been successfully applied in computer vision for classification, object detection, semantic segmentation, and visual object tracking. In addition to AlexNet, other deeper CNN networks such as VGG Net (Simonyan & Zisserman, 2014), Google Net (Szegedy et al., 2015), Deep Residual Net (He, Zhang, Ren, & Sun, 2016), DenseNet (Huang, Liu, Van Der Maaten, & Weinberger, 2017) and MobileNet (Sandler, Howard, Zhu, Zhmoginov, & Chen, 2018) have been developed.

CNNs have been successfully applied in civil engineering applications for image classifications. These include metal surface defects detection (Soukup & Huber-Mörk, 2014), post-disaster collapse classification (Yeum, Dyke, Ramirez, & Benes, 2016), joint damage detection through a one-dimensional CNN (Abdeljaber, Avci, Kiranyaz, Gabbouj, & Inman, 2017), concrete crack detection using a sliding window technique (Cha, Choi, & Büyüköztürk, 2017), pavement crack detection (Zhang et al., 2017; Vetrivel, Gerke, Kerle, Nex, & Vosselman, 2018), structural damage detection with feature extracted from low-level sensor data (Lin, Nie, & Ma, 2017), structural damage classification with the proposal of Structural ImageNet (Gao & Mosalam, 2018). Apart from the CNN-based classification, other powerful classification algorithms such as Enhanced Probabilistic Neural Network with Local Decision Circles (EPNN) and the new Neural Dynamic Classification (NDC) were successfully developed in recent years (Ahmadlou & Adeli, 2010; Rafiei & Adeli, 2017). The noticeable applications of such recent algorithms in civil engineering can be found in damage detection in high-rise buildings using neural dynamics classification (Rafiei & Adeli, 2017), development of earthquake early warning system (Rafiei, M.H. and Adeli, H. (2017), structural reliability analysis (Dai, 2017), estimation of concrete properties (Rafiei, Khushefati, Demirboga & Adeli, 2017), the global and local assessment of structural health condition using unsupervised deep Boltzmann machine (Rafiei & Adeli, 2018) and construction cost estimation (Rafiei & Adeli, 2018).

In addition to classification, CNNs can be used in the field of object detection which involves classification and localization of an object. Prior to the use of CNNs, object detection was dominated by the use of histogram of oriented gradients (HOG) (Dalal, & Triggs, 2005) and scale-invariant feature transform (SIFT) (Lowe, 2004). In 2014, Girshick, Donahue, Darrell and Malik (2014) proposed the region-based CNNs (R-CNNs), which utilizes the Region Proposal Function (RPF) in order to localize and segment objects. It significantly improved the global performance compared to the previous best result on PASCAL Visual Object Classes (VOC) challenge 2012. The PASCAL Visual VOC challenge ran each year from 2005 to 2012, which provides a benchmark in visual object category recognition and detection with a standard dataset of images and annotation, and standard evaluation procedures. Discussion of object detection proposal methods can be found in Hosang, Benenson, and Schiele (2014), Hosang, Benenson, Dollár, and Schiele (2015). Further, the Fast R-CNN (Girshick et al., 2015) and the Faster R-CNN (Ren, He, Girshick, & Sun, 2017) were developed to improve the speed and accuracy of the R-CNN. Region-based CNN methods (e.g. RCNN, Fast-RCNN, and Faster-RCNN) have been successfully implemented in civil engineering applications. Cha, Choi, Suh, Mahmoudkhani & Büyüköztürk (2018) used the Faster-RCNN to detect multiple structural damage types such as steel delamination, steel corrosion, bolt corrosion, and concrete cracks. Xue & Li, (2018) demonstrated the efficiency of Faster R-CNN for shield tunnel lining defects detection compared to traditional inspection methods. Li, Yuan, Zhang, & Yuan (2018) achieved near-real-time concrete defect detection with geolocalization using a unified vision-based methodology. Liang (2019) applied deep learning with Bayesian optimization for RC bridge damage detection.

While many previous studies can provide reasonable accuracy, they are still relatively slow in terms of achieving real-time practical application when the images were recorded with high frame per second (FPS). To address this deficiency, Redmon, Divvala, Girshick, & Farhadi, (2016) presented YOLO (i.e. You Only Look Once) for real-time object detection. While YOLO is extremely fast, it makes more localization errors and achieved relatively low recall compared to region-based CNN methods. To further improve recall and localization accuracy, Redmon & Farhadi (2017) developed the YOLOv2 algorithm. They have shown that YOLOv2 significantly improves the recall and localization accuracy while still maintaining the speed to be 10 times faster in FPS compared to Faster-RCNN on the VOC 2007 database.

CNN-based classification for civil engineering applications has been hampered by limited training data (Gao & Mosalam, 2018). In general, a single classification model can provide reasonable accuracy if the training data covers a wide range of hidden features. However, even when the size of the training data is sufficiently large, the classification model may still not perform well if the training data is not properly pre-processed to identify the localized damage (Gao & Mosalam, 2018). For example, an image may contain multiple damage states where a portion of the structure has fractured, while the other part of the structure remains undamaged.

Moreover, although region-based CNN methods have been widely applied in civil engineering, it remains almost little to none attempt of the regression-based detection methods such as YOLOv2, for structural damage detection. Therefore, the main contributions of this paper are: (a) established component training data that follow the codified damage state classification of the RC columns; (b) successfully developed and applied YOLOv2 object detection network to identify the critical damage feature of RC columns; (c) proposed and successfully implemented the dual CNN methods which incorporate the classification network and YOLOv2 object detection network to improve the accuracy achieved by a single classification network; (d) introduced performance-based assessment framework to quantify financial losses, which can be used by the decision makers for rapid emergency management and resources allocation, thus improving the regional seismic resiliency of the city.

## 2 METHODOLOGY
### 2.1 Rapid performance assessment framework

Figure 1 shows the framework proposed in this study to quantify the financial loss of the RC buildings after strong earthquake shaking. First, system-level and component-level images for a single building are collected, which can be achieved by unmanned aerial vehicles (UAVs) (Ham, Han, Lin, & Golparvar-Fard, 2016). In an ideal situation, images of the systems and components should be taken from multiple views and the most severe damage status should be considered to facilitate the comprehensive evaluation. The system-level images are assessed by a system-level classification network to confirm if the building has collapsed. If the system-level collapse is identified, the replacement cost of the building should be used. If the building is identified as non-collapse, the component-level images are fed into component-level classification and detection networks. Once the component damage states are identified, the corresponding repair costs for the components were identified from the ATC-58 (2007) fragility database. Finally, the total repair costs of the building are summed up by adding the total repair quantities from all structural and non-structural components taking into account their suitable unit cost distribution. This fragility database is one of the essential products of ATC-58 project established by the Applied Technology Council (ATC) in contracts with the Federal Emergency Management Agency (FEMA) to develop FEMA P-58 Seismic Performance Assessment of Buildings, Methodology and Implementation (also known as Performance-based Earthquake Engineering (PBEE)). Implementations of the PBEE framework for cost evaluation of buildings have been widely attempted (Goulet et al., 2007; Yang, Moehle, Stojadinovic, & Der Kiureghian, 2009;

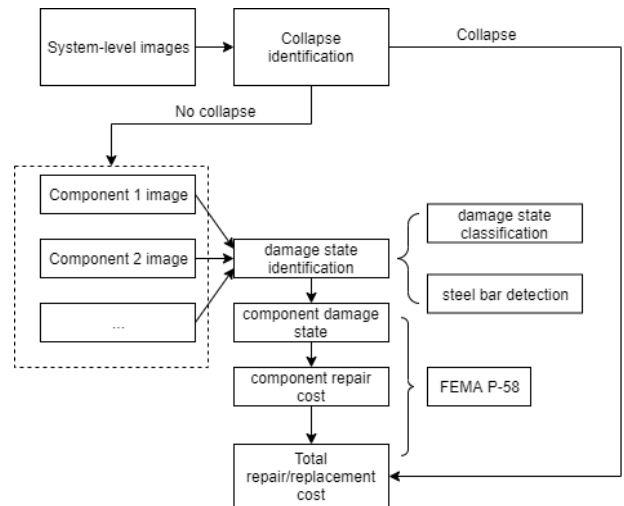

Figure 1 Flowchart of cost evaluation scheme

Mitrani-Resier, Wu, & Beck, 2016). While this study focuses on the development of dual CNN methods that employs the state-of-the-art YOLOv2 object detection algorithms to more accurately classify the structural damage state, it also makes the first attempt to integrate this fragility database with the proposed deep learning methods to facilitate cost evaluation of RC structures.

### 2.2 CNN classification

In this research, CNNs were used to identify the damage states of the building and components. Typical CNN involves multiple types of layers, including Convolution (Conv) layers, Rectified Linear Unit (ReLU) layers, Pooling layers, Fully-connected (FC) layers and Loss layer (e.g. Softmax layer). The Conv layer combined with the subsequent layer, ReLU, constitute the essential computational block for CNNs. This is the feature that distinguishes CNNs from the traditional fully connected deep learning network. One of the advantages of CNNs is that it drastically improves the computational efficiency from the traditional neural network because the number of training parameters enclosed in the filter of CNNs is significantly less than the number of weights utilized by fully connected layers which are the only layers presented in the traditional feed forward neural network. Besides, CNNs preserve the spatial locality of pixel dependencies and enforce the learnable filters to achieve the strongest response to a local input pattern.

During the forward pass, the output from the previous layer is convolved with each one of the learnable filters, which yields a stack of two-dimensional arrays. Applying the desired nonlinear activation function (such as ReLU) to these two-dimensional arrays leads to a volume of two-dimensional activation maps. After a single or multiple Conv-ReLU blocks, the pooling layer is introduced which is a form of non-linear down-sampling. The objective of the pooling layer is to reduce the number of parameters to improve the computation efficiency. During the pooling process, the input image is partitioned into sub-regions which may or may not overlap with each other. If max pooling is used, the maximum value of each sub-region is taken.

Following several Conv-ReLU blocks and pooling layers, the resulting layer is transposed to an FC layer. The output can be computed as matrix multiplication followed by a bias offset, which then substitutes into activation function. For example, in VGG-19, the CNNs end up with 3 FC layers with the dimension of 4096, 4096, and 1000, respectively. In addition, due to the fact that FC layers occupy most of the parameters in the entire CNN, they are prone to overfitting which can be alleviated by incorporating dropout layers. The idea of using dropout layers is to randomly remove FC layers, which can improve the computation efficiency and has proven to alleviate the concern for overfitting (Srivastava, Hinton, Krizhevsky, Sutskever, & Salakhutdinov, 2014). In VGG-19, 50% dropout is applied to the FC layers. Finally, the output of the last FC layer is passed to a Loss layer (i.e. Softmax in this study) which determines the probability of each class (i.e. how confidently the model thinks the input image being each class). The result of the classification is recognized for the output with the highest probability for each class.

## 2.3 System-level collapse recognition

Based on the CNNs presented, the status of the reinforced concrete building can be classified as collapse or non-collapse. Multiple pre-trained models can be used to facilitate the training process. In this study, transfer learning from the three pretrained models including AlexNet (Krizhevsky, Sutskever, and Hinton, 2012), VGG-19 (Simonyan & Zisserman, 2014) and ResNet-50 (He, Zhang, Ren, & Sun, 2016) is applied for the binary classification task. Transfer learning is a new machine learning technique that takes advantage of certain pretrained models in the source domain and fine-tunes part of the parameters with a few labeled data in the target domain, which can greatly promote the training process in the situation of data scarcity.

In deep learning, there is a trend to develop a deeper and deeper network which aims at solving more complex task and improving the performance. However, research has shown training of deep neural networks becomes difficult and the accuracy can reach plateau or even degrade (He, Zhang, Ren, & Sun, 2016). ResNets were developed by He, Zhang, Ren, & Sun, (2016) to solve the problems where the shortcut connections were proposed. It has been demonstrated that training this form of networks is easier than training plain deep convolutional neural networks and the problem of accuracy deterioration is resolved. The complete architecture of ResNet-50 adopted in this study is shown in Figure 2. The ResNet-50 contains a sequence of Convolution-Batch Normalization-ReLU (Conv-BN-ReLU) blocks. Batch normalization is added right after each convolution and before ReLU activation to stabilize training, speed up convergence, and regularize the model. After a series of Conv-BN-ReLU blocks, global average pooling (GAP) is performed to reduce the dimensionality, which is then followed by the FC layer associated with softmax function.

Due to the limited number of images for civil engineering applications, 686 images are collected from datacenterhub.org at Purdue University and google images, of which 240 images are related to the collapse of buildings, while 446 images of non-collapsed buildings. The image preprocessing is conducted to reduce the inconsistency in image classification following the same approach adopted by Gao & Mosalam (2018). The preprocessed images will be resized appropriately to 224x224 or 227x227 pixels (depending on what network is chosen) before being fed into the CNNs for state and damage classification. The performance of the model is verified through the training and validation process. In this case, 80% of the collected images are chosen as the training data and the rest is chosen as the testing data. Further, within the training set, 20% of the images are set as the validation data and the remaining images are used to train the model. Therefore, $686 \times 0.8 \times 0.8 \approx 439$, $492 \times 0.8 \times 0.2 \approx 110$, and $492 \times 0.2 \approx 137$ images are allocated for training, validation, and testing purposes, respectively.

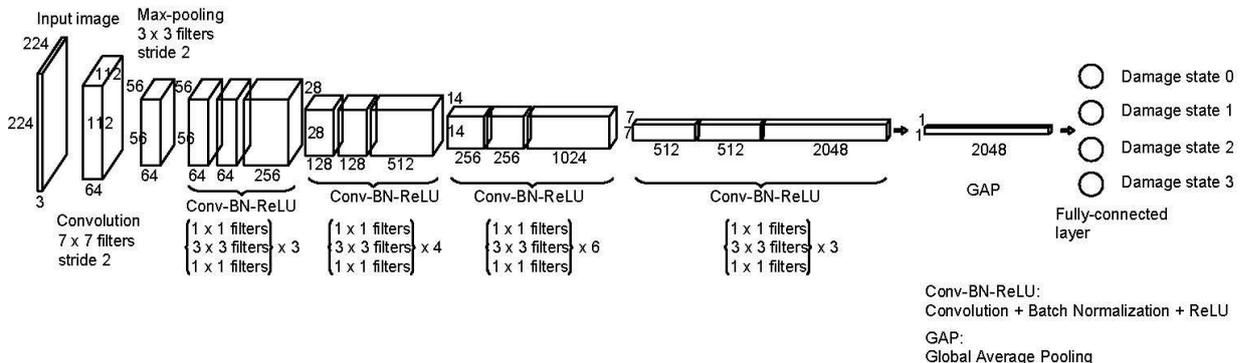

Figure 2 Architecture of ResNet-50

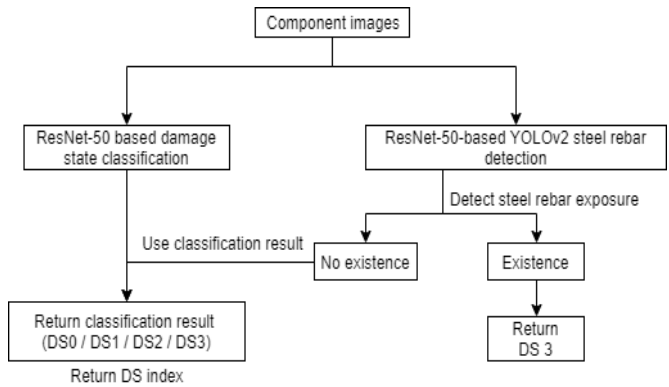

Figure 3 Flowchart of component damage state inspection scheme

## 2.4 Component-level damage state identification

As per the proposed evaluation scheme depicted in Figure 1, if the RC building is identified as non-collapse, the subsequent step is to determine the damage state of the structural components. In this study, the definition of several damage states for RC structural columns is introduced as shown in Table 1, which follows the ATC-58 damage state (DS) statement for reinforced concrete beam-column joints. Similar evaluation approaches have been shown and practically implemented for many years as demonstrated by Nakano, Maeda, Kuramoto, & Murakami, M. (2004) and

Table 1 Description of damage state classes

| DS index | Description |
|---|---|
| 0 | No damage |
| 1 | Light damage: visible narrow cracks and/or very limited spalling of concrete |
| 2 | Moderate damage: cracks, large area of spalling concrete cover without exposure of steel bars |
| 3 | Severe Damage: crushing of core concrete, and or exposed reinforcement buckling or fracture |

Maeda, Matsukawa, & Ito (2014).

The RC columns, the critical gravity supporting components of the RC buildings, are selected to demonstrate the component-level classification, detection, as well as cost evaluation. In this regard, a novel dual CNN-based inspection approach (Figure 3) is proposed to facilitate the process. On one hand, the classification model is trained across all the damage states as defined in Table 1. On the other hand, localization of steel bars is implemented using the YOLOv2 object detection approach. The advantage of the object detection approach is its ability to focus on damage-sensitive features (i.e. exposed reinforcement bars in this case) which distinguish DS 3 from DS 0, DS 1 and DS

2. It is noted that the detection of exposed reinforcement is crucial because most of the tensile stiffness and strength of the reinforced concrete components are contributed by the reinforcement. Therefore, CNN-based detection is employed to reinforce the identification of DS 3 in case the classification fails to classify it. In fact, from the safety point of view, identification of components in DS 3 condition is critical after the event of earthquake because these components are prone to fail completely in the aftershocks, which may lead to partial or complete collapse of the building and consequently significant increase of repair cost, injuries and death rate. In other words, it is more conservative to maintain the second object detection network even if in some rare cases, the final damage state is identified as DS3 while the ground truth label is one of the others.

In total, there are 2260 images collected from the damage survey conducted by Sim, Laughery, Chiou, Weng (2018), EERI Learning from Earthquake Reconnaissance Archive and Google Image. The number of images for DS 0, DS 1, DS 2 and DS 3 is 496, 404, 580 and 780, respectively. Similar as before, image preprocessing and resizing are applied before training. Also, 80% of the acquired images for each damage class is chosen as its training set and 20% as the testing set. The validation set is chosen as 20% of the training set and the rest of the training set is used to train the model.

### 2.4.1 Damage state classification

As shown in Table 1, four DS classes need to be distinguished. Similar to system-level classification, the pretrained AlexNet, VGG nets, and ResNet-50 are selected for transfer learning.

The trained model with the highest test accuracy is adopted to demonstrate the applicability of the classification of multiple damage states. The construction of the network is similar to the previous one except the last three layers, a fully connected layer, a Softmax layer, and a classification output layer are updated with new labels and the new number of classes (i.e. 4 damage states in this case).

### 2.4.2 Steel reinforcement object detection

In addition to damage state classification, a CNN-based object detection model is also introduced in this study to identify steel reinforcement exposed due to concrete spalling. In comparison to image classification, object detection is one step further which localizes the object within an image and predicts the class label of the object. The output of object detection would be different bounding boxes with their labels in the image. While R-CNN methods (i.e. R-CNN, Fast R-CNN, Faster R-CNN) have been widely attempted in civil engineering applications, they are still relatively slow for real-time applications. This study designed and applied a specific YOLOv2 object detection network for identification of reinforcement exposure. Compared to R-CNN methods, A detailed comparison of object detection networks is presented in Redmon & Farhadi (2017).

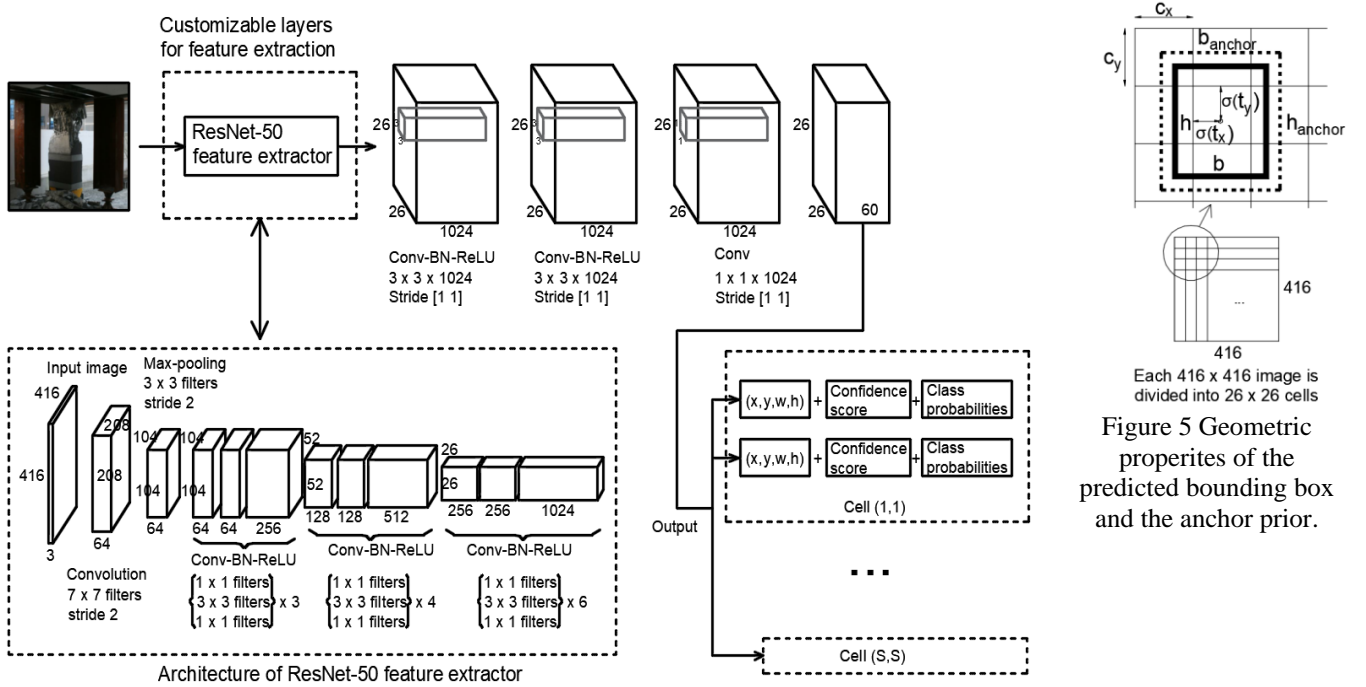

Figure 4 The schematic architecture of YOLOv2 built on ResNet-50 for steel

Figure 5 Geometric properites of the predicted bounding box and the anchor prior.

In general, YOLOv2 consists of a customized feature extractor which is usually a series of Conv-BN-ReLU blocks and pooling layers, followed by localization and classification layers which predicts the bounding box location and the class score, respectively. In this study, YOLOv2 built on ResNet-50 is adopted for steel reinforcement detection. First, the layers after the third Conv-BN-ReLU block of ResNet-50 (as shown in Figure 4) are removed such that the remaining layers can work as a feature extractor. Second, a detection subnetwork is added which comprises groups of serially connected Conv-BN-ReLU blocks. Details of layer properties within the detection subnetwork are illustrated in Figure 4. In conclusion, the detection is modelled as a regression problem. The output of the network contains S x S grid cells of which each predicts $B$ boundary boxes. Each boundary box includes 4 parameters for the position, 1 box confidence score (objectness) and $C$ class probabilities. The final prediction is expressed as a tensor with the size of $S \times S \times B \times (4 + 1 + C)$.

The objective of training of neural network is to minimize the multi-part loss function as shown in Equation (1) where $I_{i,j}^{obj} = 1$ if the $jth$ boundary box in cell $i$ is responsible for detecting the object, otherwise 0. Similarly, $I_i^{obj} = 1$ if an object appears in cell $i$, otherwise 0, and $I_{i,j}^{noobj}$ is the complement of $I_{i,j}^{obj}$. The parameters $x_i$ and $y_i$ are the predicted bounding box position, $\hat{x}_i$ and $\hat{y}_i$ refer to the ground truth position, $w_i$ and $h_i$ are the width and height of the predicted bounding box, while the associated ground truth is denoted as $\hat{w}_i$ and $\hat{h}_i$. The term $C$ is the confidence score and $\hat{C}$ is the intersection over union of the predicted bounding box with the ground truth. The multiplier $\lambda_{coord}$ is the weight for the loss in the boundary box coordinates and $\lambda_{noobj}$ is the weight for the loss in the background. As most boxes generated do not contain any objects, indicating the model detects background more frequently than detecting objects, to put more emphasis on the boundary box accuracy, $\lambda_{coord}$ is set to 5 by default and $\lambda_{noobj}$ is chosen as 0.5 by default.

$$\lambda_{coord} \sum_{i=0}^{s^2} \sum_{j=0}^{B} I_{i,j}^{obj}[(x_i - \hat{x}_i)^2 + (y_i - \hat{y}_i)^2] + \lambda_{coord} \sum_{i=0}^{s^2} \sum_{j=0}^{B} I_{i,j}^{obj}\left[\left(\sqrt{w_i} - \sqrt{\hat{w}_i}\right)^2 + \left(\sqrt{h_i} - \sqrt{\hat{h}_i}\right)^2\right] + \sum_{i=0}^{s^2} \sum_{j=0}^{B} I_{i,j}^{obj}(C_i - \hat{C}_i)^2 + \lambda_{noobj} \sum_{i=0}^{s^2} \sum_{j=0}^{B} I_{i,j}^{noobj}(C_i - \hat{C}_i)^2 + \sum_{i=0}^{s^2} I_i^{obj} \sum_{c \in classes} (p_i(c) - \hat{p}_i(c))^2 \quad (1)$$

The network learns to adapt predicted boxes appropriately with regards to ground truth data during training. However, it would be much easier for the network to learn if better anchor priors are selected. Therefore, to facilitate the training process, K-means clustering as suggested by Redmon & Farhadi (2017) is implemented to search the tradeoff between the complexity of the model and the number of bounding boxes required to achieve the desired performance. Once the number of anchors is specified, the K-means clustering algorithm takes as input the dimensions of ground truth boxes labelled in the training data, and outputs the desired dimensions of anchor boxes and the mean IoU with the ground truth data. Clearly, the selection of more anchor boxes provides higher mean IoU, but also causes more computational cost. Through the parametric study on the number of anchors, the relationship between the mean Intersection-over-Union (IoU) and the number of anchors is established in Figure 6, which shows the number of 10 anchors is a reasonable choice, where the mean IoU can reach about 0.8. It should be noted that unlike the original work by Redmon & Farhadi (2017) where the network utilizes 5 box priors to classify and localize 20 different classes, this study only focuses on the detection of one class (i.e. steel reinforcement), indicating more anchors can be used without losing too much computational efficiency. Figure 7 depicts the dimensional properties of each ground truth box as well as its associated clustering. Anchor dimensions corresponding to each cluster, determined by K-means clustering approach are reported in Table 2. These anchors will be utilized to determine the bounding box properties as shown in equations (2) to (6). In summary, $B$ is chosen as 10. $C$ is equal to 1 which corresponds to steel exposure. As a result, the predicted tensor has the size of $26 \times 26 \times 60$.

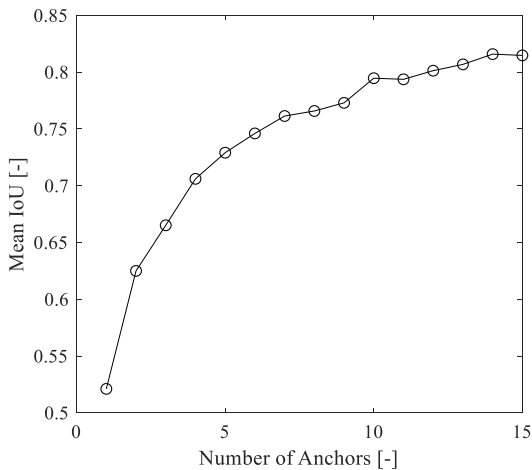

Figure 6 Relationship between Mean IoU and number of dimension priors

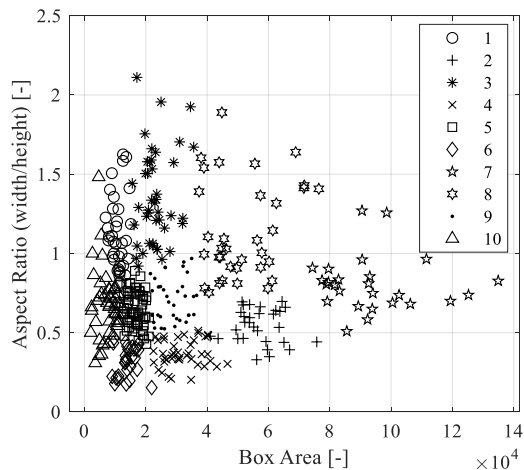

Figure 7 Illustration of K-means clustering results with 10 anchor sizes

Table 2 Selection of Anchor properties

| Group | 1 | 2 | 3 | 4 | 5 | 6 | 7 | 8 | 9 | 10 |
|---|---|---|---|---|---|---|---|---|---|---|
| Width [pixels] | 104 | 174 | 174 | 107 | 105 | 67 | 274 | 208 | 138 | 54 |
| Height [pixels] | 98 | 309 | 132 | 285 | 167 | 206 | 338 | 213 | 199 | 77 |

The network predicts 10 bounding boxes at each cell in the output feature map. For each bounding box, 5 coordinates are predicted including $t_x$, $t_y$, $t_w$, $t_h$ and $t_0$. As illustrated in Figure 5, the bold solid box is the predicted boundary box and the dotted rectangle is the anchor prior. Assuming the cell is offset from the top left corner of the image by $(c_x, c_y)$ and the anchor box prior has a width of $b_{anchor}$ and height of $h_{anchor}$, then equations (2) to (6) can be derived. Equations (2)-(3) predict the location of the bounding box and (4)-(5) predict the dimensions of the bounding box based on anchor box dimensions. Equation (6) is related to objectness prediction involves the IoU of the ground truth and the proposed box, and the conditional probability of the class given that there is an object.

$$b_x = \sigma(t_x) + c_x \qquad (2)$$
$$b_y = \sigma(t_y) + c_y \qquad (3)$$
$$b = b_{anchor}e^{t_w} \qquad (4)$$
$$h = h_{anchor}e^{t_h} \qquad (5)$$
$$\Pr(object) * IoU(b, object) = \sigma(t_0) \qquad (6)$$

Similar to other CNN models, the YOLOv2 is trained by back-propagation and stochastic gradient descent (SGD). The learning rate is constant and set to $10^{-4}$ and mini-batch size is set to 16. The input image size is 416 x 416, which is identical to what has been adopted by Redmon & Farhadi (2017) for finetuning detection subnetwork. The training and testing images of YOLOv2 are taken separately from the DS 3 images which have been used in training and testing of the DS classification model. Data augmentation such as cropping, flipping, small rotation is applied such that the augmented images still contain the object that needs to be detected. The training is implemented in MATLAB R2019a on two computers: Alienware Aurora R8 (a Core i7-9700K @ 3.60 GHz, 16 GB DDR4 memory and 8 GB memory GeForce RTX 2070 GPU) and a Lenovo Legion Y740 (a

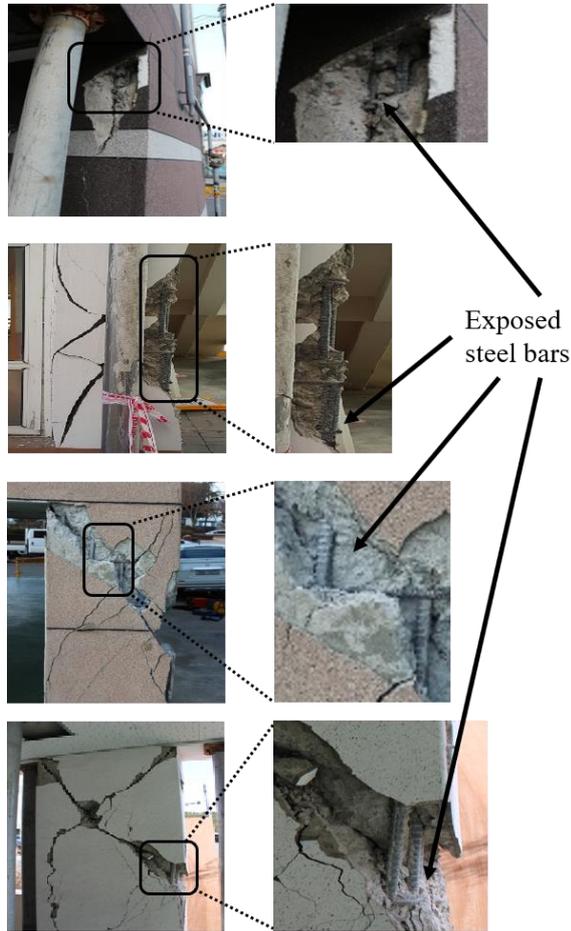

Exposed steel bars

Figure 8 Sample images of RC columns that classification model wrongly identifies as DS 2 while ground truth is DS3

Core i7-8750H @2.20 GHz, 16 GB DDR4 memory and 8 GB memory GeForce RTX 2070 max-q GPU).

### 2.4.3    Damage state determination

Post-earthquake cost evaluation of the RC building strongly relies on the classification accuracy of damage state. A single classification model generally performs well if trained on the large dataset which covers a wide range of hidden features. Besides, it is required the image scene is properly pre-processed such that the targeted region (i.e. RC column with/without damage in this study) dominates the entire image. Moreover, the classification model may not perform well if different classes have obvious shared features. In case of multiple irrelevant objects in the background or a column with multiple damage features presented in a single image (i.e. crack feature is shared by DS 1, DS2 and DS3; the spalling feature is shared by DS2 and DS3), the classification model may fail to identify the damage state class correctly. For example, an image of column is shown in Figure 8 that shows a lot of small cracks and small spalling, and at the same time also presents exposure of evident steel bars concentrated in relatively small regions, the classification model fails to identify it as DS 3 (i.e. ground truth label) in our experiments. This is interpretable because in this case, the damage features of DS 1, DS 2 and DS 3 are all included in one single image, leading to the fact that the predicted probability for DS 3 is not the highest. This is the moment when detection of steel bars is needed to reinforce the severe damage state identification for DS 3 where the exposed reinforcement needs to be specifically captured.

In this study, the results of the classification model including the label and its associated probability are obtained. Meanwhile, the reinforcement detection model checks the existence of exposed steel bars. The final decision on the damage state is taking advantage of both outcomes from the classification model and object detection model. As shown in Figure 3, each image is evaluated by the two models in parallel. For a given image, it is first resized to fit the size of input layers of the classification networks and object detection networks, respectively. The classification model predicts the probability of being each damage state and takes the one with the highest probability as the result. Meanwhile, the object detection model aims at detecting the exposed steel bars. If the steel bars are not detected, the classification result is directly output as the final inspection outcome. If the steel bars are captured by the detection model, then DS 3 should be returned as the final decision. The proposed dual CNN-based framework builds on the traditional damage classification model, and extends the evaluation scope by analyzing the local details (i.e. steel bars in this case). The object detection model does not change the fundamental diagnosis logic of the classification model but reinforces the identification of DS 3 through localization of exposed bars on top of the classification model. This comes from three bases. First, object detection is a more complex task than classification because it involves both localizing and classifying the target in the scene. Solely relying on object detection results is more likely to lower the overall evaluation accuracy (potentially caused by insufficient recall). Second, there are some other damage features of DS 3 such as crushing of concrete, substantial shear failure mechanism while the proposed object detection model is trained to detect exposed steel bars only. The identification of such multiple features still partly relies on the classification model. Third, steel reinforcement is one of the most essential load-carrying components in RC columns. The proposed detection network of steel bars introduces one redundancy to reinforce the identification of the most severe damage case which is more likely to cause an unexpected system failure in response to aftershocks, and consequently higher chances of injuries and death, as well as substantial repair cost and longer repair time.

### 2.5 Cost evaluation framework

Once the damage state is identified, the repair costs can be calculated using the fragility data presented in the FEMA P-

58 project (ATC-58, 2007). Table 3 shows an example of the damage state and repair cost for the RC column specified in FEMA P-58 project. The dispersion is defined as the standard deviation of the logarithmic value of the cost (i.e. $\sigma$).

Table 3 Mean value and dispersion of repair cost of reinforced concrete column (e.g. component ID: B1041.031a) at each damage state (ATC-58, 2007)

| Damage state index | Mean Cost (USD$) at minimum quantity | Mean Cost (USD$) at maximum quantity | Dispersion |
|---|---|---|---|
| DS 0 | 0 | 0 | 0 |
| DS 1 | 25704 | 20910 | 0.39 |
| DS 2 | 38978 | 25986 | 0.32 |
| DS 3 | 47978 | 31986 | 0.3 |

FEMA P-58 divided all vulnerable structural components, nonstructural components into fragility groups and performance groups. A fragility group is defined as a group of components, that have similarity in construction and installation techniques, modes of damage, probability of inducing damage modes as well as damage consequences. Besides, a performance group is a subset of fragility group components that are subjected to the same earthquake demands such as story drift or acceleration at a specific floor, in a specific direction.

The consequence function for repair cost as defined by FEMA P-58 is shown in Figure 9. The minimum cost refers to the unit cost to conduct a repair action, considering all possible economies of scale (which corresponds to maximum quantity) and operation efficiencies. On the contrary, the maximum cost is the unit cost with no benefits from scale and operation efficiencies, which corresponds to the minimum quantity. If needed, unit repair costs uncertainties can be accounted for using normal or lognormal distribution.

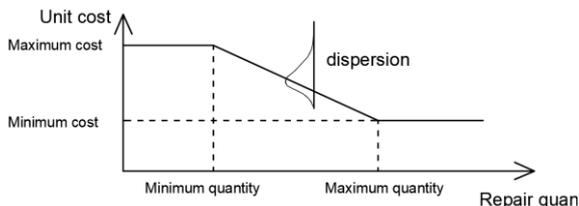

Figure 9 Typical consequence function for repair costs

In this study, the cost evaluation scheme is inspired by part of the seismic performance assessment methodology from FEMA P-58. To initiate the proposed cost evaluation scheme, the global images of the building should be assessed by the global classification model to identify whether the building experiences collapse. If collapse occurs, the replacement cost of the building is reported. If the building does not have global failure, the damage state inspection approach is continued. In this case, the images of components are processed by the local damage classification model and object detection model in parallel to identify the damage state. The cost of the component is retrieved based on the damage state identified. Finally, the total repair cost of the building is calculated as the sum of each performance group which contains structural or non-structural components.

$$Total\ repair\ cost = \sum_{i}^{l} \sum_{j}^{m} \sum_{k}^{n} C_{i,j,k}\left(\sim logN(\mu, \sigma^2)\right) \quad (7)$$

Where $l$ denotes the number of fragility groups, $m$ denotes the number of performance groups and $n$ denotes the number of components within each performance group. The term $C_{i,j,k}$ is the repair cost drawn from the normal or lognormal distribution based on the identified damage state. Finally, the Monte Carlo process is repeated in accordance with the assumed unit cost distribution for a large number of realizations, of which each represents one total loss value. These realizations are sorted in ascending order and a lognormal distribution is fitted to facilitate the calculation of the probability that total loss will be less than any specific value. The results are presented in a loss curve which can be used for risk management decisions.

## 3 EXPERIMENTS AND RESULTS
### 3.1 System-level failure classification

This subsection presents the results from three different pretrained models as shown in Table 4. Although the testing accuracy among all three pretrained models is relatively close, the ResNet-50 which has the deepest architecture, yields slightly higher accuracy than the other two CNN models. The loss and accuracy of ResNet-50 during training are presented in Figure 10. Both the training and validation accuracy exceeds 90% after around 80 epochs of training and approaches 100% at the end. It is generally acknowledged that the validation dataset can provide an unbiased assessment of a model fit on the training dataset (Krizhevsky, Sutskever, and Hinton, 2012; He, Zhang, Ren, & Sun, 2016). Increase in the error on the validation dataset is a sign of overfitting to the training dataset. A high and stable validation accuracy of the validation is observed in Figure 10 which demonstrates the applicability of the system-level classification model for collapse identification. Figure 11 compares the confusion matrices (Kohavi & Provost, 1998) between training and testing results. For instance, 95% of the testing images which have ground-truth labels as collapse are successfully predicted while only 5% of these images are misclassified as "no collapse". Moreover, sample testing images with probability for its associated class are shown in Figure 12. The trained model can predict the correct class for the images with high probability.

Table 4 System-level and component-level training parameters and performance of transfer learning from three different pretrained models

| Pretrained CNN models | AlexNet | VGG-19 | ResNet-50 |
|---|---|---|---|
| Input size | $227 \times 227 \times 3$ | $224 \times 224 \times 3$ | $224 \times 224 \times 3$ |
| Initial learning rate | 0.0001 | 0.0001 | 0.0001 |
| Regularization factor | 0.0001 | 0.0001 | 0.0001 |
| Momentum coefficient | 0.90 | 0.90 | 0.90 |
| System-level testing accuracy | 93.15% | 95.63% | 95.92% |
| Component-level testing accuracy | 85.17% | 87.17% | 87.47% |

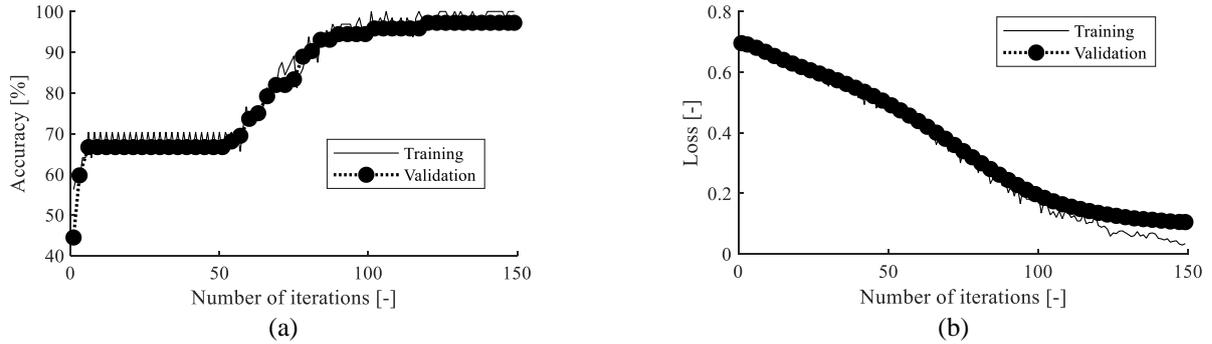

Figure 10 System-level collapse identification for training and validation sets using ResNet-50: (a) accuracy and (b) loss

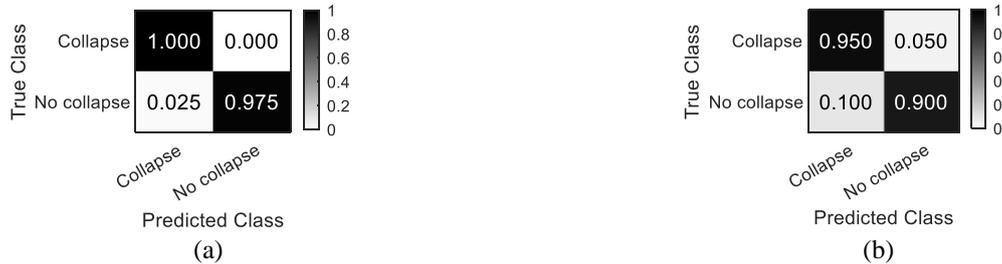

Figure 11 System-level collapse versus no collapse: confusion matrices of (a) training set and (b) testing set

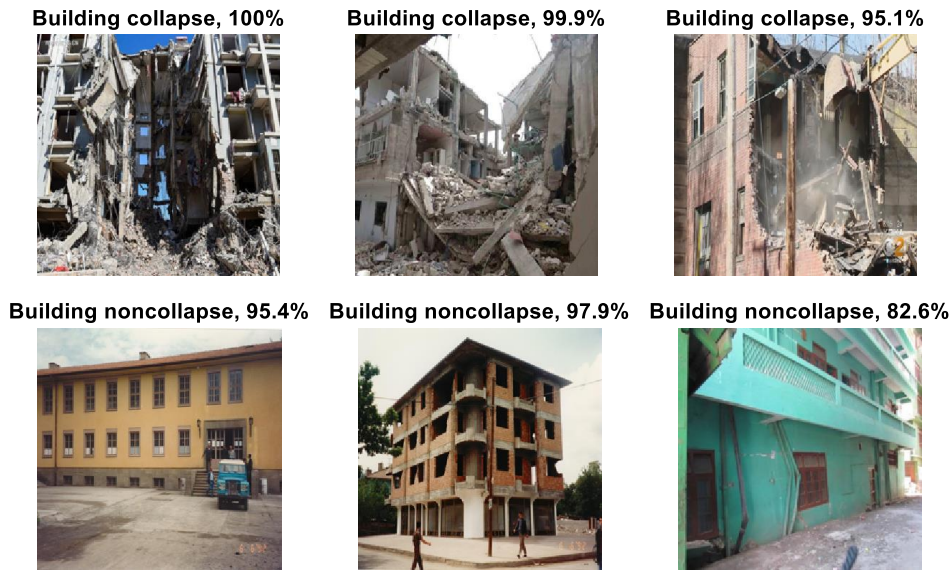

Figure 12 Sample testing images of the building with predicted probability for each class

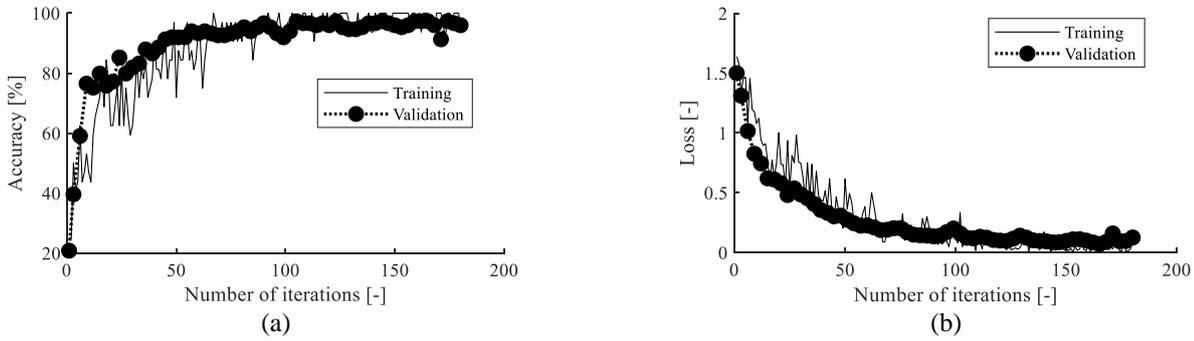

Figure 13 Component-level DS classification for training and validation sets using ResNet-50: (a) accuracy and (b) loss

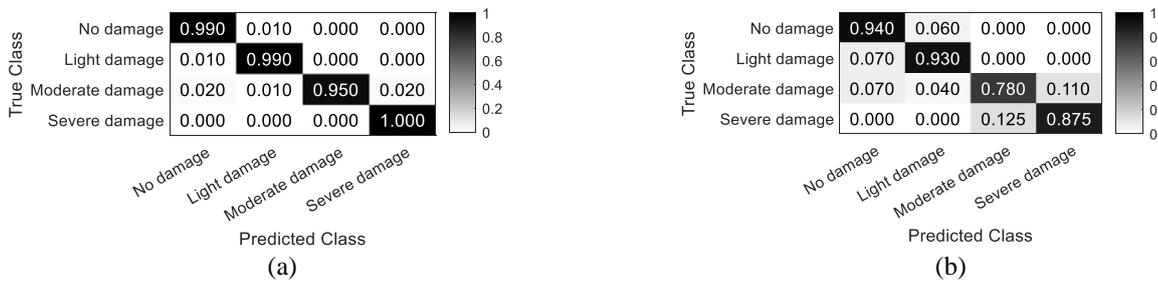

Figure 14 Component-level damage state identification: confusion matrices of (a) training (left) and (b) testing set.

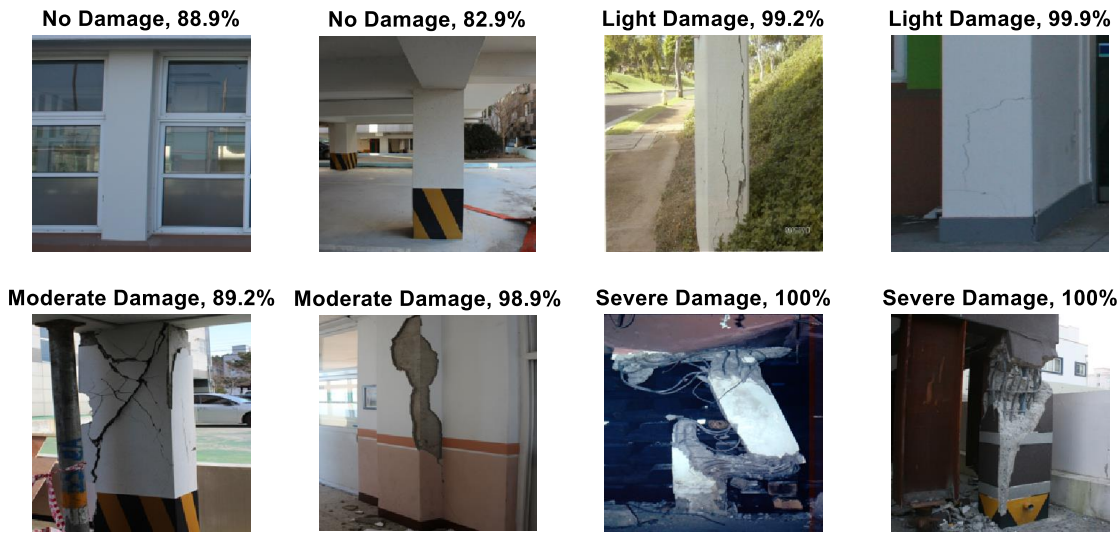

Figure 15 Reasonable prediction of sample testing images of the building with predicted probability for each class

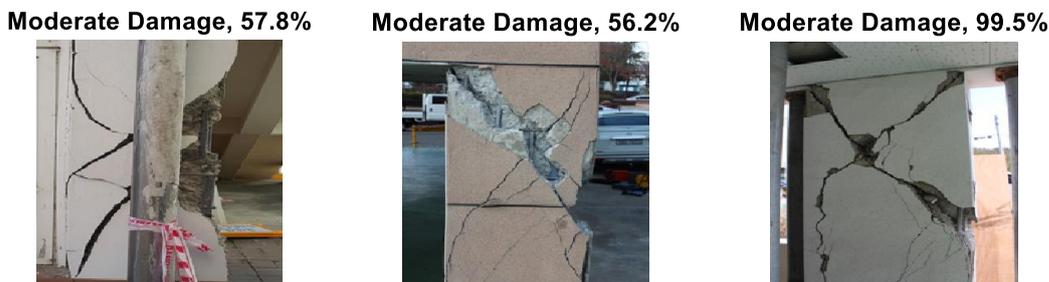

Figure 16 Unreasonable prediction of sample testing images with ground truth of "Severe Damage" as shown in Figure 8

## 3.2 Component-level damage state identification
### 3.2.1    Damage state classification

Similar to the system-level failure classification, Table 4 presents the identical training parameters and performance comparison of three different pretrained models (AlexNet, VGG-19, ResNet-50) for the classification of the component damage states. In general, all three models have high accuracy, while the ResNet-50 has slightly higher accuracy than AlexNet and VGG-19. The loss and accuracy for ResNet-50 during the training process are presented in Figure 13, which shows both the training and validation accuracy are approaching 100% at the end. The performance of the trained model is confirmed by the confusion matrix for training and testing as shown in Figure 14. Figure 15 shows the classification of a few sample images with correct prediction. The results show that the trained model is able to classify different damage states with reasonably high accuracy, although the classification accuracy with respect to moderate damage (i.e. DS 2) and severe damage (i.e. DS 3) are not as high as that regarding the class of no damage (i.e. DS 0) and light damage (i.e. DS 1). The results reflect there is an increasing difficulty in detecting damage features from DS 0 to DS 3. Basically, there is no damage feature in DS 0 when RC columns are in almost perfect condition in which case the trained CNN model only needs to identify column profile without any extra damage features. Similarly, only cracks and very limited spalling are enclosed in DS 1, where slightly more damage features are introduced compared to DS 0. However, more features are usually observed in DS 2, such as light or severe cracks and a large area of spalling. In the case of DS 3, the model performs reasonably well to detect its own damage features which include exposure of significant length of steel bars, crushing of concrete and buckling or fracture of reinforcement. However, it occasionally misclassifies the damage state as DS 2, while the ground truth is DS 3 (Figure 16). There are two potential reasons. First, DS 2 and DS 3 have many common damage features, such as cracks and a large amount of concrete spalling. Second, exposure of steel reinforcement is not evident in some cases while cracks or significant spalling may be dominant the entire image (Figure 8). To overcome such deficiency, a novel object detection technique is implemented and combined with the classification technique to identify the damage states (Figure 3). Details of the integrated method are described in Session 3.2.3.

### 3.2.2    Steel reinforcement object detection

This subsection presents the results regarding the detection of exposed longitudinal reinforcement in RC columns to demonstrate the applicability of YOLOv2 in this scenario. The performance of an object detector is usually evaluated using a precision-recall curve (Everingham, Van Gool, Williams, Winn, & Zisserman, 2010). A low false positive rate leads to high precision and low false negative rate results in a high recall. In other words, a large area under the recall-precision curve indicates the high performance of the detector with both high recall and precision. A detector with high recall but low precision retrieves many results, but most of its predicted labels are incorrect (e.g. incorrect bounding box locations, low IoU). A detector with high precision but low recall can localize the object very accurately once the object is successfully recalled, but only a very few results can be recalled. The average precision (AP) is often used to quantify the performance of an object detector (Girshick, 2015; Ren, He, Girshick, & Sun, 2017), which is determined as the area under the precision-recall curve. Mean AP (mAP) is defined as the mean of calculated APs for all classes. Figure 17 presents the precision-recall curve for training and testing. The mAP for both training and testing has demonstrated the applicability of YOLOv2 for detecting steel bars. The testing results in Figure 17 (b) indicate there still exists room for improving the mAP with the larger training set, particularly improving the recall. It should be noted that the detection of steel bars is more difficult than detecting objects with a more regular pattern because the steel bars may buckle or fracture in a very complex way in different situations. Figure 18 provides sample detection images where the steel bars are localized by rectangular bounding boxes. Figure 18(b) (upper) shows the images which were wrongly classified by the traditional classification method, while Figure 18(b) (lower) shows the same images in which the exposed steel bars are localized by YOLOv2.

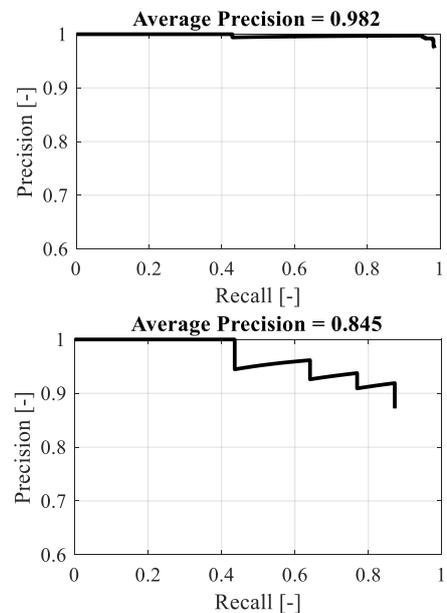

Figure 17 Recall - precision curve of training (upper) and testing (testing)

### 3.2.3    Damage state determination

As illustrated in Figure 3, the proposed evaluation scheme integrates the classification model and object detection model to reinforce the identification of the most severe damage state. Figure 19 presents the identification summary of a single classification model and the dual CNNs model using the confusion matrix method. Overall, the employment of the YOLOv2 object detector improves the identification accuracy of DS 3 by 7.5%. It should be noted that the classification accuracy can be potentially improved by expanding the training on the large dataset. However, the training data for civil engineering applications is relatively limited as aforementioned. The YOLOv2 network demonstrated in this study can be considered as a local-feature-oriented detector on the basis of existing classification networks to specially focus on the identification of DS 3 images when the training dataset is relatively small.

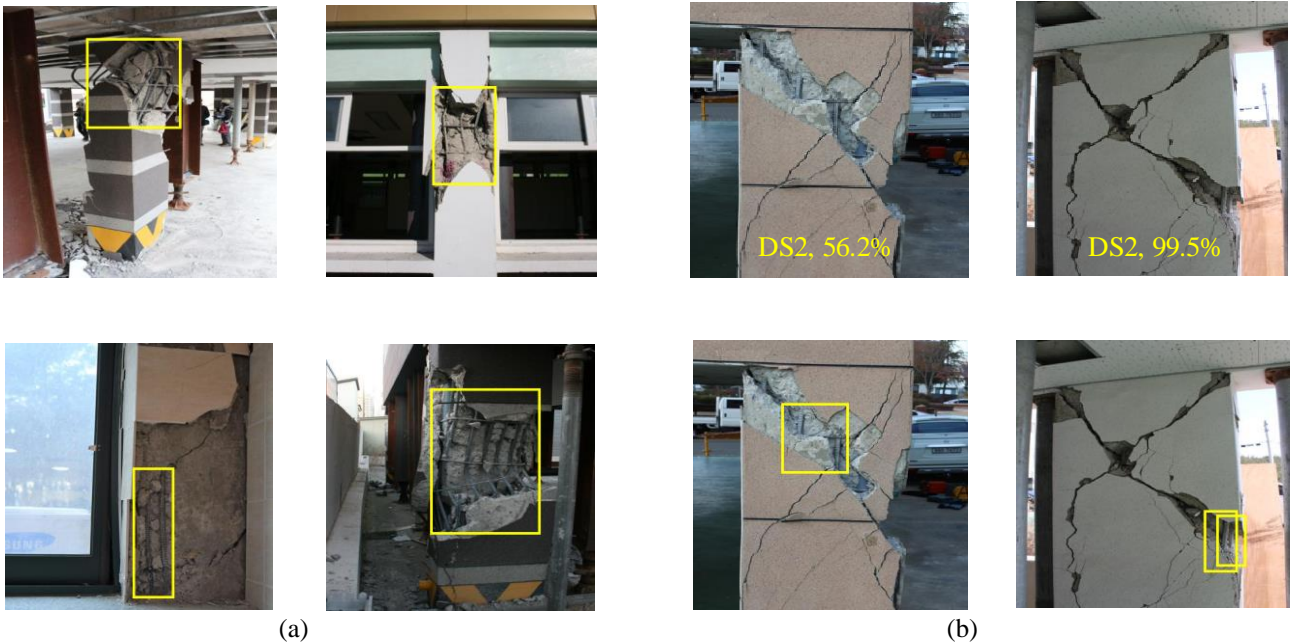

(a) (b)

Figure 18 Detection of steel bars highlighted by yellow rectangular bounding boxes (a) Sample testing images, (b) detection of exposed steel bars (lower) in testing images wrongly predicted by classification model (upper)

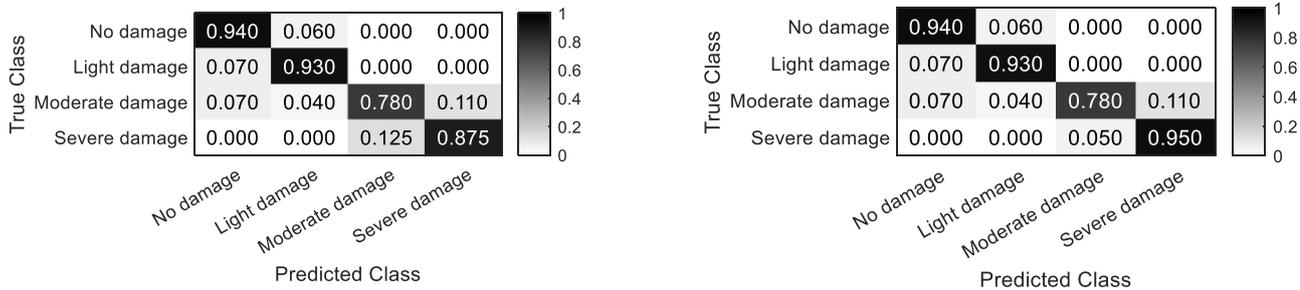

Figure 19 Confusion matrix with consideration of only classification model (left); both classification and object detection model (right)

### 3.2.4 Repair cost evaluation

The damage state determined by the proposed dual CNN algorithms can be integrated with the cost evaluation framework as presented in Section 2.5. For the purpose of illustration, a prototype RC building, as shown in Figure 20, is selected (Sim et al., 2015) to evaluate its damage and repair cost after an earthquake. In this study, the repair cost evaluation is based on the damage states of the available images collected. In summary, the dual CNNs algorithms determine that 17 of such columns are in DS1, 26 in DS2, and 14 in DS3. Using the unit cost information provided in Table 3, the total repair cost of these columns is calculated. It should be noted that only the RC columns are considered in this study but similar procedures can be applied to all components. The process is repeated 10000 times with Monte Carlo procedures to simulate the dispersion of the repair costs. Finally, the results are presented in a cumulative distribution function as shown in Figure 21. The cost simulation results can provide critical risk data for decision making and resource allocation during post-disaster reconstruction. For example, the decision maker can use the 50% probability of non-exceedance to identify the median repair cost for the building. In the example presented in Figure 21, the median repair cost is $2.69 million USD for the prototype building.

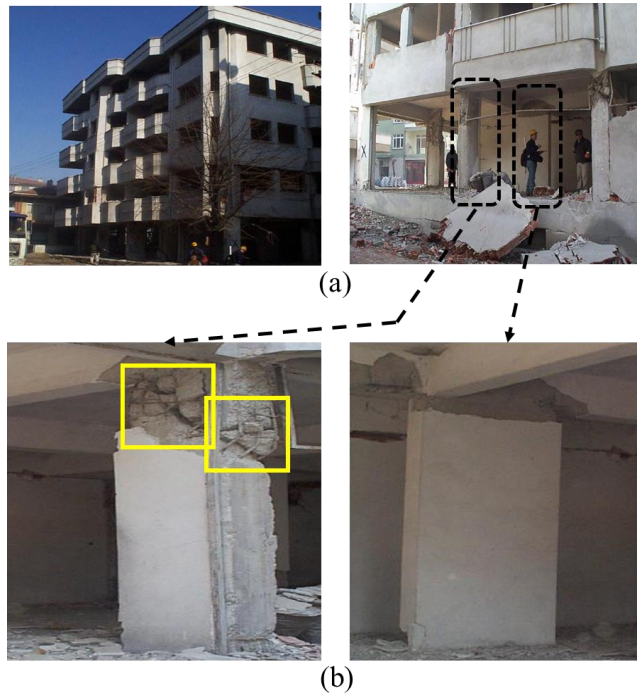

(a)

(b)

Figure 20 Case study of a RC building with sample results (a) system-level identification: non-collapse and (b) component-level damage evaluation: severe damage with detetction of steel exposure (left), and moderate damage (right)

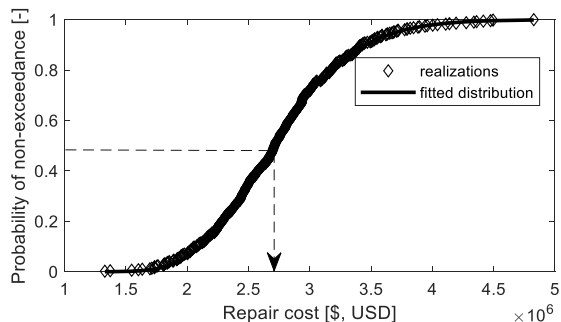

Figure 21 Repair cost distribution corresponding to the hypothetical case

## 4 CONCLUSIONS

Rapid post-disaster damage estimation and cost evaluation of RC structures are becoming a crucial need for building owners and decision makers for risk management and resource allocation. This paper proposed a rapid cost evaluation framework which incorporates the state-of-the-art image processing techniques to quantify the structural damages and the integration with financial loss estimation of RC structures. Multiple innovations are presented in this paper: 1) Both system-level and component-level classification models were trained successfully, which follows post-disaster damage state quantification guidelines of RC structures; 2) A state-of-the-art real-time object detector, named YOLOv2 built on ResNet-50 was introduced and first implemented to demonstrate its applicability for detecting exposure of steel bars with 98.2% and 84.5% mAP in both training and testing process. 3) In comparison to a single classification network, the implementation of the trained YOLOv2 combined with classification network improves the accuracy by 7.5% in identifying the most severe damage state which imposes critical threats for life safety and contributes the most to repair cost. 4) a novel integration of performance assessment framework with the damage detection methods is proposed and implemented to facilitate the repair cost evaluation, which can be easily conveyed to decision makers and stake holders who lack of engineering knowledge. Overall, the proposed framework shows the rapid inspection of the RC buildings with components is possible using image-based classification and detection techniques, there still remains room to improve the recall in steel bars detection. The concept of the dual CNN scheme and the integration with cost estimation can be considered and extrapolated by other researchers for damage detection and loss evaluation of other structural types such as masonry, steel and timber structures.

## 5 ACKNOWLEDGMENTS

The authors would like to acknowledge the funding provided by the International Joint Research Laboratory of Earthquake Engineering (ILEE), National Natural Science Foundation of China (grant number: 51778486), Natural Sciences and Engineering Research Council (NSERC), China Scholarship Council. Any opinions, findings, and conclusions or recommendations expressed in this paper are those of the authors.

## 6 REFERENCES

Abdeljaber, O., Avci, O., Kiranyaz, S., Gabbouj, M., & Inman, D. J. (2017). Real-time vibration-based structural damage detection using one-dimensional convolutional neural networks. *Journal of Sound and Vibration*, 388, 154–170.

Ahmadlou, M., & Adeli, H. (2010). Enhanced probabilistic neural network with local decision circles: A robust classifier. *Integrated Computer-Aided Engineering*, 17(3), 197-210.

ATC-58. (2007). "Development of Next Generation Performance-Based Seismic Design Procedures for New and Existing Buildings", ATC, Redwood City, California, U.S.A. https://www.atcouncil.org/p-58

Cha, Y. J., Choi, W., & Büyüköztürk, O. (2017). Deep learningbased crack damage detection using convolutional neural networks. *Computer-Aided Civil and Infrastructure Engineering*, 32(5), 361–378.

Cha, Y. J., Choi, W., Suh, G., Mahmoudkhani, S., & Büyüköztürk, O. (2018). Autonomous structural visual inspection using region-based deep learning for detecting multiple damage types. *Computer-Aided Civil and Infrastructure Engineering*, 33, 9–11.

Chungwook Sim; Cheng Song; Nick Skok; Ayhan Irfanoglu; Santiago Pujol; Mete Sozen (2015), "Database of low-rise reinforced concrete buildings with earthquake damage,"

Dai, H. (2017). A wavelet support vector machine-based neural network meta model for structural reliability assessment. *Computer-Aided Civil and Infrastructure Engineering*, 32, 4, 344–357.

Dalal, N., & Triggs, B. (2005). Histograms of oriented gradients for human detection. *In international Conference on computer vision & Pattern Recognition (CVPR'05)*, 1, 886-893, IEEE Computer Society.

Everingham, M., Van Gool, L., Williams, C. K., Winn, J., & Zisserman, A. (2010). The pascal visual object classes (voc) challenge. *International journal of computer vision*, 88(2), 303-338.

Gao, Y, & Mosalam, K. M. (2018). Deep transfer learning for imagebased structural damage recognition. *Computer-Aided Civil and Infrastructure Engineering*, 33, 372–388.

Girshick, R. (2015). Fast R-CNN. *In Proceedings of the IEEE International Conference on Computer Vision*, 1440–1448.

Girshick, R., Donahue, J., Darrell, T., & Malik, J. (2014). Rich feature hierarchies for accurate object detection and semantic segmentation. *In Proceedings of the IEEE conference on computer vision and pattern recognition*, 580-587.

Goulet, C. A., Haselton, C. B., Mitrani-Reiser, J., Beck, J. L., Deierlein, G. G., Porter, K. A., & Stewart, J. P. (2007). Evaluation of the seismic performance of a code-conforming reinforced-concrete frame building—from seismic hazard to collapse safety and economic losses. *Earthquake Engineering & Structural Dynamics*, 36(13), 1973-1997.

Ham, Y., Han, K. K., Lin, J. J., & Golparvar-Fard, M. (2016). Visual monitoring of civil infrastructure systems via camera-equipped Unmanned Aerial Vehicles (UAVs): a review of related works. *Visualization in Engineering*, 4(1), 1.

He, K., Zhang, X., Ren, S., & Sun, J. (2016). Deep residual learning for image recognition. *In Proceedings of the IEEE Conference on Computer Vision and Pattern Recognition*, 770–778.

Hinton, G. E., Osindero, S., and Teh, Y. (2006). A fast learning algorithm for deep belief nets. *Neural Computation*, 18, 1527–1554. 14, 19, 27, 143, 528, 529, 660, 661

Hosang, J., Benenson, R., & Schiele, B. (2014). How good are detection proposals, really? Presented at *Proceedings of British Machine Vision Conference*, Nottingham, England.

Hosang, J., Benenson, R., & Dollár, P., & Schiele, B. (2015). What makes for effective detection proposals? *IEEE Transactions on Pattern Analysis and Machine Intelligence*, 38(4), 814–830.

Hoskere, V., Park, J. W., Yoon, H., & Spencer Jr, B. F. (2019). Vision-Based Modal Survey of Civil Infrastructure Using Unmanned Aerial Vehicles. *Journal of Structural Engineering*, 145(7), 04019062.

Huang, G., Liu, Z., Van Der Maaten, L., & Weinberger, K. Q. (2017). Densely connected convolutional networks. *In Proceedings of the IEEE conference on computer vision and pattern recognition*, 4700-4708.

Jahanshahi,M. R., Kelly, J. S.,Masri, S. F., & Sukhatme, G. S. (2009). A survey and evaluation of promising approaches for automatic imagebased defect detection of bridge structures. *Structure and Infrastructure Engineering*, 5(6), 455–486.

Japan Building Disaster Prevention Association. (1991). Guideline for post-earthquake damage evaluation and rehabilitation.

Koch, C., Georgieva, K., Kasireddy, V., Akinci, B., & Fieguth, P. (2015). A review on computer vision based defect detection and condition assessment of concrete and asphalt civil infrastructure. *Advanced Engineering Informatics*, 29(2), 196–210.

Kohavi, R., & Provost, F. (1998). Confusion matrix. *Machine Learning*, 30(2-3), 271–274.

Krizhevsky, A., Sutskever, I., & Hinton, G. E. (2012). ImageNet classification with deep convolutional neural networks. *In Advances in Neural Information Processing Systems*, 1097–1105.

LeCun, Y., Bottou, L., Bengio, Y., & Haffner, P. (1998). Gradient-based learning applied to document recognition. *Proceedings of the IEEE*, 86(11), 2278–2324.

Li, R., Yuan, Y., Zhang, W., & Yuan, Y. (2018). Unified vision-based methodology for simultaneous concrete defect detection and geolocalization. *Computer-Aided Civil and Infrastructure Engineering*, 33(7), 527-544.

Liang, X. (2019). Image-based post-disaster inspection of reinforced concrete bridge systems using deep learning with Bayesian optimization. *Computer-Aided Civil and Infrastructure Engineering*, 34(5), 415-430.

Lin, Y. Z., Nie, Z. H., & Ma, H. W. (2017). Structural damage detection with automatic feature-extraction through deep learning, *Computer-Aided Civil and Infrastructure Engineering*, 32, 12, 1025–1046.

Lowe, D. G. (2004). Distinctive image features from scale-invariant keypoints. *International journal of computer vision*, 60(2), 91-110.

Maeda, M., Matsukawa, K., & Ito, Y. (2014, July). Revision of guideline for post-earthquake damage evaluation of RC buildings in Japan. *In Tenth US national conference on* *earthquake engineering, frontiers of earthquake engineering*. Anchorage, Alaska, 21-25.

McCulloch, W. S. and Pitts, W. (1943). A logical calculus of ideas immanent in nervous activity. *Bulletin of Mathematical Biophysics*, 5, 115–133. 14, 15.

Mitrani-Resier, J., Wu, S., & Beck, J. L. (2016). Virtual Inspector and its application to immediate pre-event and post-event earthquake loss and safety assessment of buildings. *Natural Hazards*, 81(3), 1861-1878..

Nakano, Y., Maeda, M., Kuramoto, H., & Murakami, M. (2004, August). Guideline for post-earthquake damage evaluation and rehabilitation of RC buildings in Japan. *In 13th World Conference on Earthquake Engineering* (No. 124).

Rafiei, M. H., & Adeli, H. (2017). A new neural dynamic classification algorithm. *IEEE transactions on neural networks and learning systems*, 28(12), 3074-3083.

Rafiei, M. H., & Adeli, H. (2017). A novel machine learning-based algorithm to detect damage in high-rise building structures. *The Structural Design of Tall and Special Buildings*, 26(18), e1400.

Rafiei, M. H., & Adeli, H. (2017). NEEWS: a novel earthquake early warning model using neural dynamic classification and neural dynamic optimization. *Soil Dynamics and Earthquake Engineering*, 100, 417-427.

Rafiei, M. H., & Adeli, H. (2018). A novel unsupervised deep learning model for global and local health condition assessment of structures. *Engineering Structures*, 156, 598–607.

Rafiei, M. H., Khushefati, W. H., Demirboga, R., & Adeli, H. (2017). Supervised Deep Restricted Boltzmann Machine for Estimation of Concrete. *ACI Materials Journal*, 114(2).

Rafiei, M.H. and Adeli, H. (2018), "A Novel Machine Learning Model for Construction Cost Estimation Taking Into account Economic Variables and Indices," *Journal of Construction Engineering and Management*, 144:12, 2018, 04018106 (9 pages)

Redmon, J., & Farhadi, A. (2017). YOLO9000: better, faster, stronger. *In Proceedings of the IEEE conference on computer vision and pattern recognition*, 7263-7271.

Redmon, J., Divvala, S., Girshick, R., & Farhadi, A. (2016). You only look once: Unified, real-time object detection. *In Proceedings of the IEEE conference on computer vision and pattern recognition*, 779-788.

Ren, S., He, K., Girshick, R., & Sun, J. (2017). Faster R-CNN: Towards real-time object detection with region proposal networks. *IEEE Transactions on Pattern Analysis and Machine Intelligence*, 39(6), 1137–1149.

Rosenblatt, F. (1958). The perceptron: A probabilistic model for information storage and organization in the brain. *Psychological Review*, 65, 386–408. 14, 15, 27.

Rumelhart, D., Hinton, G., and Williams, R. (1986). Learning representations by back-propagating errors. *Nature*, 323, 533–536. 14, 18, 23, 204, 225, 373, 476, 482.

Sandler, M., Howard, A., Zhu, M., Zhmoginov, A., & Chen, L. C. (2018). Mobilenetv2: Inverted residuals and linear bottlenecks. *In Proceedings of the IEEE Conference on Computer Vision and Pattern Recognition* (pp. 4510-4520).

Sim, C., Laughery, L., Chiou, T. C., Weng, P. (2018), "2017 Pohang Earthquake - Reinforced Concrete Building Damage Survey," https://datacenterhub.org/resources/14728.

Simonyan, K., & Zisserman, A. (2014). Very deep convolutional networks for large-scale image recognition. *arXiv preprint* arXiv:1409.1556.

Soukup, D., & Huber-Mörk, R. (2014). Convolutional neural networks for steel surface defect detection from photometric stereo images. *International Symposium Visual Computing*, NewYork, NY: Springer International Publishing, 668–677.

Szegedy, C., Liu, W., Jia, Y., Sermanet, P., Reed, S., Anguelov, D., … Rabinovich, A. (2015). Going deeper with convolutions. *In Proceedings of the IEEE Conference on Computer Vision and Pattern Recognition*, 1–9.

Vetrivel, A., Gerke, M., Kerle, N., Nex, F. C., & Vosselman, G. (2018). Disaster damage detection through synergistic use of deep learning and 3Dpoint cloud features derived fromvery high resolution oblique aerial images, and multiple-kernel-learning, *ISPRS Journal of Photogrammetry and Remote Sensing*, 140, 45–59.

Xue, Y. D., & Li, Y. C. (2018). A fast detection method via region-based fully convolutional neural networks for shield tunnel lining defects. *Computer-Aided Civil and Infrastructure Engineering*, 33, 8.

Yang, T. Y., Moehle, J., Stojadinovic, B., & Der Kiureghian, A. (2009). Seismic performance evaluation of facilities: methodology and implementation. *Journal of Structural Engineering*, 135(10), 1146-1154.

Yeum, C. M., & Dyke, S. J. (2015). Vision-based automated crack detection for bridge inspection. *Computer-Aided Civil and Infrastructure Engineering*, 30, 10, 759-770.

Yeum, C. M., Dyke, S. J., Ramirez, L., & Benes, B. (2016). Big visual data analysis for damage evaluation in civil engineering. *In International Conference on Smart Infrastructure and Construction*, Cambridge, U.K., June 27–29.

Zeiler, M. D., & Fergus, R. (2014, September). Visualizing and understanding convolutional networks. *In European conference on computer vision*, 818-833. springer, Cham.

Zhang, A., Wang, K. C. P., Li, B., Yang, E., Dai, X., Peng, Y., … Chen, C. (2017). Automated pixel-level pavement crack detection on 3D asphalt surfaces using a deep-learning network. *Computer-Aided Civil and Infrastructure Engineering*, 32(10), 805–819.

Zhou, B., Khosla, A., Lapedriza, A., Oliva, A., & Torralba, A. (2016). Learning deep features for discriminative localization. *In Proceedings of the IEEE conference on computer vision and pattern recognition*, 2921-2929.

Zhu, Z., & Brilakis, I. (2010). Concrete column recognition in images and videos. *Journal of Computing in Civil Engineering*, 24(6), 478– 487.